# Multi-temporal Adaptive Red-Green-Blue and Long-Wave Infrared Fusion for You Only Look Once-Based Landmine Detection from Unmanned Aerial Systems

James E. Gallagher[1], Edward J. Oughton[1], Jana Košecká[2]

[1] Department of Geography & Geoinformation Science, George Mason University, Fairfax, VA 22030, USA.
[2] Department of Computer Science, George Mason University, Fairfax, VA 22030, USA.

Corresponding author's email: jgalla5@gmu.edu

Contributing authors: eoughton@gmu.edu; kosecka@gmu.edu;
These authors contributed equally to this work.

## Abstract

Landmines remain a persistent humanitarian threat, with 110 million actively deployed mines across 60 countries, claiming 26,000 casualties annually. This research evaluates adaptive Red-Green-Blue (RGB) and Long-Wave Infrared (LWIR) fusion for Unmanned Aerial Systems (UAS)-based detection of surface-laid landmines, leveraging the thermal contrast between the ordnance and the surrounding soil to enhance feature extraction. Using You Only Look Once (YOLO) architectures (v8, v10, v11) across 114 test images, generating 35,640 model-condition evaluations, YOLOv11 achieved optimal performance (86.8% mAP@0.5), with 10–30% thermal fusion at 5–10m altitude identified as the optimal detection parameters. A complementary architectural comparison revealed that while RF-DETR achieved the highest accuracy (69.2% mAP@0.5), followed by Faster R-CNN (67.6%), YOLOv11 (64.2%), and RetinaNet (50.2%), YOLOv11 trained 17.7 times faster than the transformer-based RF-DETR (41 minutes versus 12 hours), presenting a critical accuracy-efficiency tradeoff for operational deployment. Aggregated multi-temporal training datasets outperformed season-specific approaches by 1.8–9.6%, suggesting that models benefit from exposure to various thermal conditions. Anti-Tank (AT) mines achieved 61.9% detection accuracy, compared with 19.2% for Anti-Personnel (AP) mines, reflecting both the size differential and thermal-mass differences between these ordnance classes. As this research examined surface-laid mines where thermal contrast is maximized, future research should quantify thermal contrast effects for mines buried at varying depths across heterogeneous soil types.

## 1 INTRODUCTION

Landmines constitute one of the most persistent and devastating post-conflict humanitarian challenges, with an estimated 110 million landmines dispersed across 60 countries, remaining active for decades after hostilities end[1–3]. Each year, approximately 26,000 people are killed or injured by landmines, with children representing 42% of these victims[4]. Beyond the human toll, the economic and social costs are also profound, rendering vast tracts of potentially productive land unusable and impeding post-conflict recovery and development[5,6]. Furthermore, traditional landmine detection methods remain hazardous, inefficient, and prohibitively expensive[7,8]. Using existing demining methods would require centuries to clear landmines worldwide. To make matters worse, the ongoing conflict in Ukraine has resulted in Russian forces laying extensive mine fields throughout the country, intensifying the urgency for innovative landmine detection solutions[9,10].

When integrated effectively, several emerging technologies offer promising avenues for revolutionizing landmine detection. First, the rapid advancement of Unmanned Aerial Systems (UAS) has enabled landmine detection from safer distances, with significantly broader coverage than traditional ground-based methods[11,12]. UAS platforms also support various optical and electromagnetic detection systems, providing tailored detection solutions[13]. Second, deep learning models, particularly computer vision algorithms, have demonstrated remarkable improvements in speed, accuracy, and edge deployment capabilities in recent years[14,15]. Although there are several Convolutional Neural Networks (CNNs) to choose from, the You Only Look Once (YOLO) algorithm stands out among them.

YOLO represents a revolutionary approach to object detection that frames the detection problem as a single regression task, simultaneously predicting bounding boxes and class probabilities in one forward pass through the neural network[16,17]. Unlike traditional two-stage detection methods that first generate region proposals and then classify them, YOLO's unified architecture enables real-time inference speeds while maintaining accuracy, making it suitable for time-critical applications such as demining operations[18,19]. YOLO is used in several applications that require real-time detection, including military, security, and search and rescue operations[20–24]. Furthermore, YOLO's continuous evolution through successive versions has introduced architectural improvements that address specific detection challenges, including enhanced small object detection capabilities crucial for identifying mines, as well as improved feature extraction mechanisms[25–28]. Implementing machine learning solutions for landmine detection can substantially reduce human labor, decrease time requirements, and significantly improve safety for demining personnel[29,30]. Third, the increasing adoption of Long-Wave Infrared (LWIR) sensors for computer vision applications enhances the viability of thermal sensing for landmine detection[31,32].

Unlike conventional Red-Green-Blue (RGB) sensors, LWIR sensors excel at detecting surface-laid landmines that accumulate heat during daylight hours[33,34]. Thermal infrared imaging systems effectively detect landmines by capturing the

## Landmines Used in this Research Study

An assortment of globally deployed Anti-Personnel (AP) and Anti-Tank (AT) landmines with metal and plastic casings.

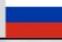

Figure 1. Anti-Personnel (AP) and Anti-Tank (AT) landmines used for this research, consisting of 10x AT plastic, 2x AT metal, 3x AP metal, and 6x AP plastic mines.

distinct thermal signatures created as these objects transfer heat at different rates than the surrounding soil[35,36]. This thermal contrast becomes most pronounced during temperature transition periods, when the thermal radiance emitted by the landmines contrasts with the cold ground[37,38]. In Ukraine, drone operators have begun employing UAS equipped with thermal cameras and other complementary sensors to identify landmines (Fig. 2)[39,40]. Ironically, UAS are also being used to emplace mines in Ukraine[41]. However, while thermal imaging can detect landmine presence, determining landmine type based solely on thermal signatures remains challenging due to lower resolution in LWIR images[42,43]. Conversely, RGB sensors, despite limitations in initial landmine detection across complex terrains, provide superior image resolution for landmine classification once detected.

This research addresses critical gaps in the literature on multispectral landmine detection[44]. We integrate the detection capabilities of LWIR with the classification strengths of RGB to enhance both detection and identification of various landmine types, ranging from larger Anti-Tank (AT) mines to smaller Anti-Personnel (AP) mines. Furthermore, this study measures multispectral object detection performance in identifying both metal and plastic landmines across AT and AP categories. We also address the research gap concerning UAS-based landmine detection utilizing fusion-enabled RGB and LWIR sensor payloads[44]. This study quantifies the performance of adaptable multispectral fusion, in which RGB and LWIR integration is optimized based on environmental conditions, including illumination and temperature variables. While this study does not implement real-time adaptive sensor fusion, we evaluate adaptive multispectral performance using post-flight data. To assess the effectiveness of adaptive multispectral fusion in identifying surface-laid landmines, this study addresses three primary research questions:

1. What is the optimal RGB-LWIR fusion approach when evaluated against environmental variables such as air temperature, ground temperature, and illumination?

2. What are the optimal RGB-LWIR fusion ratios for detecting land mines given mine characteristics and sensor altitude?

3. How do transformer-based, two-stage, and YOLO architectures perform when processing adaptive multispectral landmine imagery?

This research provides key scientific insights into the effectiveness of adaptable multispectral fusion for UAS-based object detection in support of humanitarian demining operations. Lastly, this research contributes a novel multispectral dataset, which we call the Adaptive Multispectral Landmine Identification Dataset (AMLID)[45]. AMLID consists of 12,078 images, with incremental fusion levels of 10%, and includes 21 common landmines used worldwide. The subsequent section presents a comprehensive literature review, followed by methodological details in Section 3. Section 4 presents empirical results, while Section 5 discusses findings and implications. Finally, Section 6 offers concluding observations and recommendations for future research directions.

## 2 Literature Review

The development of computer vision-based landmine detection systems has seen rapid advances, particularly with the integration of deep learning algorithms and UAS. Recent research has demonstrated the superior performance of object detection models in identifying surface-laid landmines across different environmental conditions[46–50]. Comparative studies reveal significant variation in detection accuracy across different neural network architectures, with a comprehensive evaluation of four computer vision foundation models: YOLOF, DETR, Sparse-RCNN, and VFNet. In the study, YOLOF achieved the highest performance with a mAP of 89% for drone imagery captured at 10 meters above ground level[46]. Exceptional results have also been achieved using YOLOv8, reporting a precision of 92.9%, a recall of 84.3%, and a Mean Average Precision (mAP) of 93.2% on a dataset of 1,055 landmine images captured under various lighting and environmental conditions[51].

The increasing demand for real-time object detection capabilities has also led to a growing use of YOLO model applications for landmine detection[52,53]. One study used a YOLOv8 demining robot that achieved two frames per second processing while missing only 1.6% of targets for PFM-1 (butterfly) and PMA-2 (starfish) surface landmines[48]. Another study created a comprehensive workflow that combined YOLO with geoinformation systems, enabling complete integration from detector training to coordinate extraction for operational use[54].

While YOLO's single-stage architecture excels at real-time detection, several other computer vision architectures in the literature are used for multispectral fusion applications. Popular two-stage detectors, such as Faster R-CNN, employ region proposal networks followed by refinement stages, enabling improved localization, which is crucial for detecting small objects such as AP mines at higher altitudes where resolution is degraded[55–57]. Another group of computer vision architectures that are gaining in popularity are transformer-based models, such as RF-DETR, which leverages global self-attention mechanisms that can dynamically weight RGB and LWIR features, offering potential advantages for adaptive fusion by automatically emphasizing one sensor modality over another under varying illumination conditions[58–60]. Additionally, RetinaNet is a model that addresses class imbalance between foreground and background by using a Feature Pyramid Network (FPN) backbone for multi-scale feature representation, combined with focal loss[61]. This method assigns higher weights to challenging object classes while reducing the impact of easily classified examples, making it particularly effective for imbalanced datasets common in multispectral applications[61]. Understanding the comparative performance of these distinct models relative to contemporary YOLO models under varying environmental conditions and fusion configurations is essential for developing operationally effective multispectral landmine detection systems.

The literature also expounds on the compelling use case of LWIR sensors as an effective detection method when compared to visible-spectrum imaging, particularly for buried or camouflaged landmines[62]. MobileNetV3-Large architecture for thermal drone imagery analysis achieves a training mAP of 96.1% across 2,700 thermographic images[63]. Another study achieved successful results in detecting AP landmines through thermographic imaging, demonstrating a 97.1% mAP detection accuracy for images acquired at 1 meter altitude and 88.8% at higher altitudes using Multilayer Perceptron classifiers [64].

The literature also discusses how environmental factors significantly impact detection performance, necessitating robust algorithmic approaches. Vegetation coverage poses a particular challenge, with recent modeling demonstrating that detection accuracy decreases as plant cover increases, following mathematical patterns that help determine the best times and locations for UAV collection[65]. Landmine detection challenges are increasingly being addressed through dual-mode fusion approaches, combining visible light and infrared imagery to achieve 97.1% mAP detection accuracy and 60 frames per second (FPS)[66].

## 2.1 Multispectral and Hyperspectral Landmine Detection

While landmine detection methods demonstrate general success with RGB imagery, multispectral and hyperspectral imaging technologies offer unique advantages for landmine detection by exploiting spectral signatures that may not be visible in conventional imaging systems[67,68]. These approaches leverage the distinct reflectance characteristics of landmines and surrounding materials across multiple spectral bands[69,70]. These techniques compare pixel intensities across multiple wavelengths to detect anomalous spectral signatures characteristic of explosive materials, demonstrating excellent detection performance with reasonable computational complexity when combined with computer vision models[71]. Hyperspectral sensors are also excellent at identifying landmines in complex environments, such as dense vegetation[72]. However, the trade-off with hyperspectral sensors is that they are not real-time and require substantial time and computational power to generate results. Advanced hyperspectral modeling techniques have achieved high-precision spatial resolution for landmine detection, capturing targets at sub-pixel level precision ($0.945 \times 0.945$ mm) within broader scene areas of $1.868 \times 1.868$ cm, resulting in individual landmines being represented by 52-116 pixels for detailed analysis[73].

Modern multispectral approaches increasingly emphasize sensor fusion to overcome limitations of individual sensing modalities. Detection-driven fusion architectures integrate multispectral data with YOLOv5 detection networks, specifically addressing object occlusion challenges in scatterable landmine detection[74]. For example, joint training algorithms enable semantic information to flow dynamically into fusion networks, improving recall rates, particularly for occluded targets. Decision-level data fusion approaches have also been developed, which combine multispectral and infrared imagery from UAV platforms, utilizing statistical methods to improve landmine detection accuracy[75]. Furthermore, a systematic comparison of traditional classifiers against convolutional neural networks for multispectral landmine detection achieved an above 97% overall accuracy in both indoor and outdoor environments[69]. This research reveals that deep learning approaches show increased performance for larger mines but expected decreased effectiveness for smaller land mines. The study also highlighted the importance of soil composition in detection performance.

## 2.2 Other Landmine Detection Approaches from Autonomous Systems

Beyond optical and multispectral detection methods, autonomous systems can employ various sensor configurations and platform variants to address the complex challenge of landmine detection across various operational environments[76]. For example, magnetometers represent a well-established approach for metallic landmine identification[77]. Drones equipped with magnetometer sensors have undergone successful field tests worldwide, establishing fundamental design parameters, including optimal flight speed and altitude for magnetometer-integrated UAS[78–82]. Magnetometer applications have also been extended through deep learning integration with airborne magnetometry imaging and edge computing, demonstrating the viability of combining traditional geophysical methods with modern computational approaches[83].

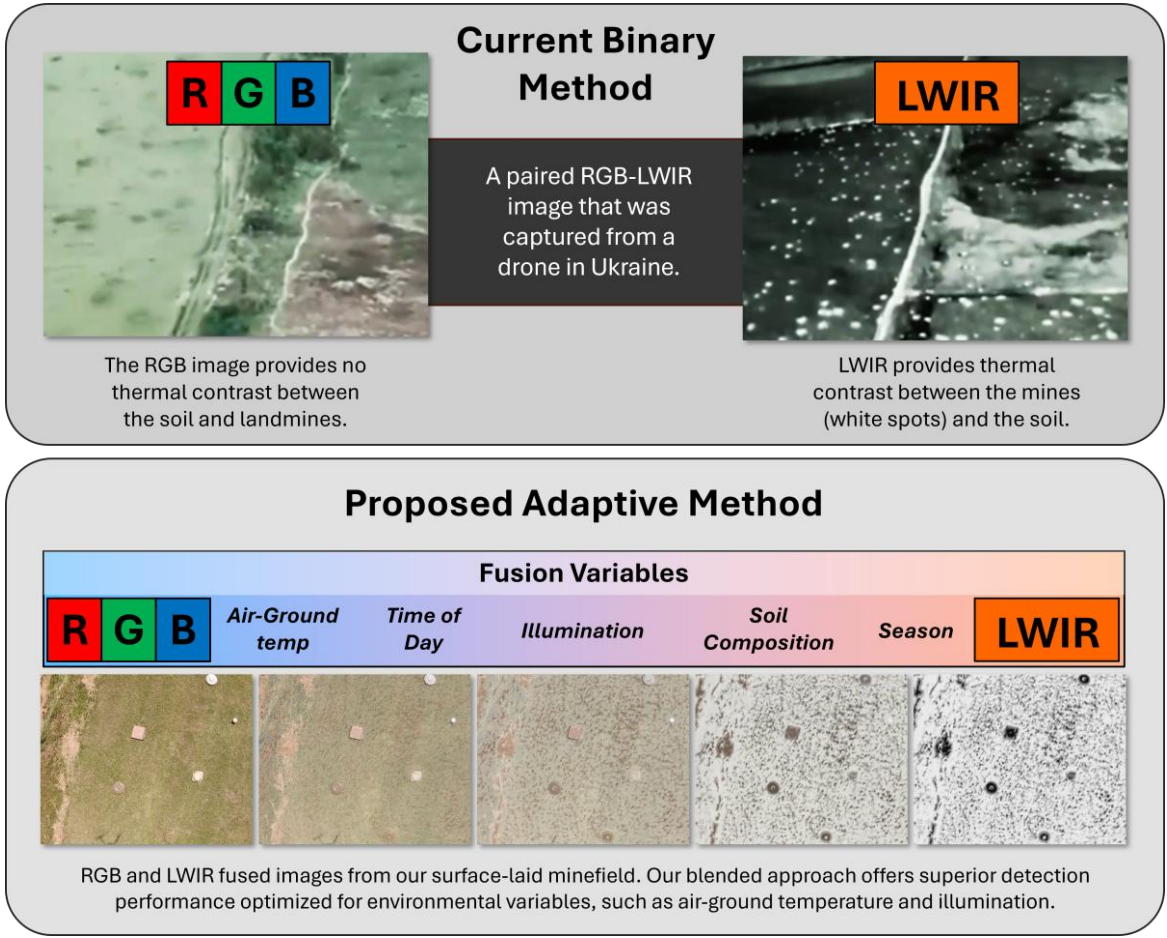

Figure 2. Conventional binary approaches require operators to select between RGB or LWIR modalities, whereas the proposed adaptive method optimizes fusion ratios along the RGB-LWIR continuum based on environmental variables including air-ground temperature differential, time of day, illumination, soil composition, and season.

Subsurface detection capabilities are essential for identifying buried landmines, particularly when surface-based methods prove insufficient. Two key systems for sub-surface landmine detection are Ground Penetrating Radar (GPR) and metal detectors[84,85],[86–88]. For example, the integration of YOLOv8 with GPR images achieved notable advances in subsurface object detection with a mAP of 93% for contemporary plastic landmines that exhibit low reflectivity[89]. There have also been several other YOLO use cases with GPR and metal detection integration[90–93]. Furthermore, there is also a growing use of autonomous systems with multiple sensing modalities for comprehensive detection and demining capabilities[94–98]. One such example is a multi-sensor dataset featuring dual-view sensor scans from an unmanned ground vehicle with a robotic arm that integrated two LiDAR sensors with images captured across RGB, Short Wave Infrared (SWIR), and LWIR[99]. This provides multiple viewpoints to mitigate occlusions and improve spatial awareness through approximately 38,000 RGB frames, 53,000 VIS-SWIR frames, and 108,000 LWIR frames[99]. High-resolution radar imaging is another capability that offers a unique method for landmine detection through advanced signal processing techniques[100–102]. Another approach is Constant False Alarm Rate Networks (CFARNets), which represent an innovative approach designed explicitly for ultrawideband synthetic aperture radar applications[91]. These networks transform traditional CFAR detectors into trainable filters that can learn to identify landmines while maintaining the interpretable features that make CFAR detection reliable. CFARNets achieve detection performance comparable to state-of-the-art methods while offering faster processing speeds[103].

## 3 Method

The method section will be broken down into the following steps: (1) data collection and labeling, (2) model selection, (3) model training, and (4) model testing and evaluation.

### 3.1 Data Collection

This research used 21 inert mines from the U.S. Army's Counter Explosive Hazards Center (CEHC). These mines can be categorized into four groups. The first group is AT and AP mines. AT mines are three to six times larger in both dimensions and weight since they are designed to penetrate armored vehicles. In contrast, AP mines are meant to be used against personnel and are much smaller and harder to detect with human and machine methods. The second category between AP and AT mines is metal and plastic mines. Metal mines are not widely used in contemporary conflicts because they are easily detected with magnetometers and are more susceptible to corrosion than plastic mines. Modern mines are composed mainly of plastic, since plastic is considerably harder to detect with metal-detecting technology and is also resistant to corrosion. Figure 1 displays all the mines used in this research, consisting of ten plastic AT mines, two AT metal mines, three AP metal mines, and six AP plastic mines.

The RGB-LWIR fusion method employed in this study uses feature alignment, with alpha adjustments applied to the LWIR layer, while the RGB layer receives no adjustments. The LWIR sensor used to collect landmine data was the FLIR Vue Pro R. This radiometric camera had a 45° field of view (FOV) and a 6.8 mm lens diameter. The LWIR resolution is 336 × 256 pixels, with a spectral band of 7.5–13.5 μm. The RGB sensor selected was the RunCam 5, which uses a Sony IMX377 12-megapixel image sensor with a 145° FOV and a resolution of 1920 × 1080 pixels. A 3D-printed housing was designed and printed to bolt the cameras as close together as possible to reduce parallax.

Data was collected in January and May in Leavenworth, Kansas. The average air temperature for January in this location was 16.5 C, while the average air temperature in May was 29.8 C, resulting in a 13.3 C difference between the two collection months. Ground temperature in January was 0.13 C, and in May it was 29.76 C, a difference of 29.63 C between the two months. Table I

# Method for Multispectral Landmine Detection
This method outlines data collection, model training, and sensor evaluation for AT and AP with multiple models and variables.

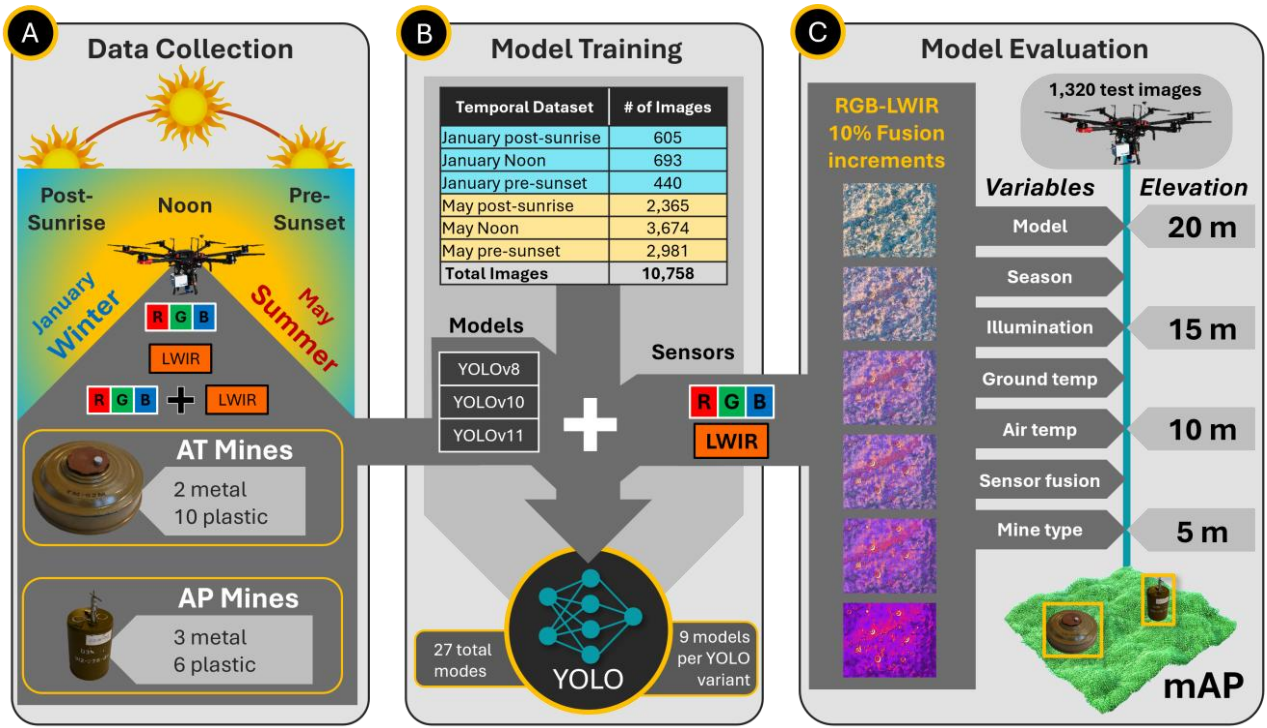

Figure 3. Method visualized with (A) Data Collection, (B) Model Training, and (C) Model Evaluation steps.

provides a detailed breakdown of temperature and illumination conditions during the month-specific time period. Data was collected during three time periods: post-sunrise, noon, and pre-sunset. These times were selected to measure the performance of adaptable multispectral fusion under varying illumination and temperature conditions. The post-sunrise period had the coldest average ground and air temperatures (5.2 C and 16.3 C), with low illumination (7,608 lux). The noon period is characterized by high ground and air temperatures (21 C and 24 C) and very high illumination (59,950 lux), with minimal shadows. Lastly, pre-sunset has mid-range ground and air temperatures (18.6 C and 21.8 C) and medium-level illumination (18,399 lux) with high cast shadows. Data was collected using a DJI Inspire 2 at fixed elevations ranging from 5 m to 30 m.

Following data collection, footage was fused post-flight, with RGB footage as the base layer and LWIR footage overlaid using feature alignment. The LWIR alpha level was adjusted in 10% increments, with 0% alpha corresponding to a pure RGB image and 100% alpha resulting in full thermal input. This fusion level is shown in Figure 3C. This systematic data collection approach can be mathematically represented in formula 1 as the union of all possible combinations across the three primary experimental dimensions. The comprehensive dataset structure encompasses every intersection of temporal conditions (month and time of day) with multispectral fusion parameters, ensuring complete coverage of the environmental spectral parameter space. This approach yields 66 distinct condition combinations, providing a robust experimental framework for evaluating the performance of adaptive multispectral landmine detection.

$$(1) \quad \cup_{(m \in M)} \cup_{(t \in T)} \cup_{(a \in A)} D_{m,t,a} = D_{total}$$

The variable *M* represents the two collection months of January and May, capturing seasonal environmental variations. The temporal dimension *T* encompasses the three daily collection periods of post-sunrise, noon, and pre-sunset, designed to measure performance across varying illumination and temperature conditions. The fusion parameter *A* comprises 11 incremental levels from 0.0 to 1.0 in steps of 0.1, where each level represents the alpha blending weight applied to the LWIR layer while leaving the RGB base layer unchanged. Each dataset subset $D_{m,t,a}$, therefore, contains all images collected under the specific combination of month (*m*), time period (*t*), and fusion level (*α*)

Following the footage fusion, frames were extracted every 2 seconds (120 frames) for each fusion level. Including pure RGB and LWIR inputs, there were 11 fusion levels. Images were then manually analyzed to remove blurry or poor-quality images. Following the quality-control review of the images, all landmine images were manually labeled. Because landmines are such an uncommon object class, automated labeling techniques that use the COCO dataset yielded poor results when attempting to label them automatically. Following labeling, the dataset comprises 12,078 images (10,758 for training and 1,320 for testing). Figure 3 provides a breakdown of the images by month and time period. Due to collection constraints, there is an imbalance between the January and May collection periods. January accounts for 16.2% of the training dataset (1,738 images), while May accounts for 83.8% (9,020 images).

### 3.2 Model Selection

This study employs three of the newest YOLO architectures: YOLOv8, YOLOv10, and YOLOv11, which represent the most recent advances in real-time object detection. The first model evaluated, YOLOv8, was developed by Ultralytics as the direct successor to YOLOv5, with significant architectural improvements[104]. Building upon YOLOv5's foundation, YOLOv8 introduces key innovations, including the replacement of the C3 module with the more efficient C2f module, the adoption of an anchor-free split head design (departing from YOLOv5's anchor-based approach), and enhanced versatility across multiple computer vision tasks, including pose estimation, instance segmentation, and oriented object detection. These architectural refinements provide optimized trade-offs between accuracy and speed while maintaining the proven reliability of the YOLOv5 lineage, making it suitable for real-time landmine detection applications.

TABLE I
Ground Temperature, Air Temperature, and Illuminance Measurements During January and May in Leavenworth, Kansas

| Variable | January | | | May | | | Average |
|---|---|---|---|---|---|---|---|
| | Post-Sunrise | Noon | Pre-Sunset | Post-Sunrise | Noon | Pre-Sunset | |
| Ground Temp (C°) | -10.9 | 8.5 | 2.8 | 21.4 | 33.5 | 34.4 | 14.9 |
| Air Temp (C°) | 9.9 | 20.7 | 18.9 | 22.6 | 27.4 | 24.7 | 20.7 |
| Lux (lx) | 3,716 | 36,300 | 5,498 | 11,500 | 83,600 | 31,300 | 28,652 |

Table 1. Environmental variables for data collection periods.

The next model, YOLOv10, was developed by researchers at Tsinghua University, and represents a paradigm shift in object detection by eliminating Non-Maximum Suppression (NMS) during inference through consistent dual assignments for training[105]. YOLOv10 architecture incorporates lightweight classification heads, spatial channel decoupled down sampling, and rank-guided block design to reduce computational redundancy. YOLOv10-S achieves 1.8× faster performance than RT-DETR-R18 while maintaining similar AP on COCO, indicating significant efficiency improvements suitable for landmine detection applications.

The last YOLO algorithm evaluated is YOLOv11, which introduces enhanced architectural innovations, including the C3k2 (Cross Stage Partial with kernel size 2) block for improved feature extraction and C2PSA (Convolutional block with Parallel Spatial Attention) components that enable more nuanced detail capture while maintaining real-time inference capabilities[106]. The model implements Spatial Pyramid Pooling–Fast (SPPF) for efficient multiscale feature aggregation and refined training methodologies that enhance both speed and accuracy. YOLOv11-N achieves competitive performance with reduced parameter count compared to its predecessors, representing a significant advancement in computational efficiency while maintaining state-of-the-art detection accuracy[107].

Lastly, to comprehensively evaluate YOLO's performance against non-YOLO models, this study extends the architectural comparison to include three models that represent distinct design philosophies. These models are RF-DETR (transformer-based detection), Faster R-CNN (two-stage detection), and RetinaNet (single-stage detection using focal loss to mitigate class imbalance). This expanded evaluation enables a systematic assessment of how the best-performing YOLO model (YOLOv11) from this research performs against distinct object detection architectures on the AMLID dataset. All four models will be trained using 15 epochs. The complete training implementation and comparative analysis code can be found in the Google Colab notebook in the data section, which provides a comprehensive evaluation of all four architectures across 1,320 test images spanning four altitudes, six temporal periods, and eleven fusion levels.

**3.3 Model Training**

This model training strategy systematically explores the complete parameter space defined by the intersection of YOLO architectures and temporal datasets, ensuring that each architectural variant is evaluated across all environmental conditions. The training configuration space mathematically represents the 27 distinct models, each processing 11 fusion levels across six temporal conditions.

$$(2) \quad \{\theta_{ij} : i \in \{Y8, Y10, Y12\}, j \in \{D_1, D_2, \ldots, D_9\}\} = \theta$$

$\theta$ represents the complete model training configuration space encompassing all possible combinations of YOLO architecture and temporal datasets, while the architecture dimension $i$ includes the three YOLO variants *Y8, Y10, and Y11*. The dataset dimension $j$ encompasses the nine temporal datasets $D$ derived from the combination of two collection months with three daily time periods. Each configuration $\theta_{i,j}$ represents a unique trained model instance, yielding three architectures and nine temporal datasets, resulting in 27 distinct models that collectively evaluate 264 unique fusion-temporal-elevation combinations across 7,128 total model-condition evaluations.

Each model configuration undergoes standardized training protocols to ensure comparative validity across architectures. The training implementation includes 50 epochs and early stopping after 20 epochs if mAP does not increase to prevent overfitting. A standard 640×640-pixel image resolution and a batch size of 16 were used during model training. The preprocessing pipeline converts Pascal VOC XML annotations to YOLO format using normalized bounding box coordinates. Data augmentation techniques include mosaic augmentation, random horizontal flipping, brightness, and contrast adjustments to enhance model generalization across different environmental conditions. Each temporal dataset comprises 11 fusion levels, ranging from 0% to 100% LWIR alpha blending with RGB base layers, resulting in a comprehensive multispectral representation. Model training used the standard 80/20 train/validation split, followed by model initialization with pre-trained weights to accelerate transfer learning. Key performance metrics were extracted, including mAP@50, precision, recall, and inference timing.

For a comprehensive architectural comparison, the three non-YOLO models and the best-performing YOLO model (YOLOv11, RF-DETR, Faster R-CNN, and RetinaNet) were trained on the same AMLID training dataset with an 80/20 training/validation split. Given the extended training time required for transformer-based architectures, all models were trained for only 15 epochs using architecture-specific optimizers (training RF-DETR for 15 epochs took 12 hours). Training was conducted using the same 1,320-image test set across all four architectures to ensure consistent evaluation metrics. This multi-model training strategy will enable a performance comparison among YOLO's single-stage anchor-free approach, transformer-based global attention mechanisms, classical two-stage region proposal refinement, and focal loss-based class imbalance handling across the full spectrum of multispectral fusion conditions and environmental variables.

### 3.4 Model Testing and Evaluation

Model performance evaluation employs standard object detection metrics consistent with contemporary computer vision research, enabling direct comparison with existing landmine detection literature and broader object detection benchmarks. mAP at IoU 0.5 (mAP@50) as the primary performance indicator measuring detection accuracy with 50% Intersection over Union threshold (IoU). Model evaluation will compare the performance of YOLOv8, YOLOv10, YOLOv11, RF-DETR, Faster R-CNN, and RetinaNet across seasons, time of day, and fusion-level ratios. mAP results will be automatically aggregated into a CSV file, which will be used to generate plots and findings to identify the top-performing dataset and model configurations across different environments to detect landmines.

This comprehensive approach ensures robust statistical validation of findings while identifying optimal configurations for model and sensor fusion rate for system deployments. The testing dataset used consists of 1,320 test images that were not used for training or validation. The testing dataset maintains balanced representation across all operational conditions, with each of the six temporal periods (January post-sunrise, January noon, January pre-sunset, May post-sunrise, May noon, May pre-sunset) contributing 220 images distributed equally across four elevation levels (5m, 10m, 15m, 20m) with 55 images per elevation-temporal combination. This symmetrical distribution ensures comprehensive evaluation across the full range of environment and altitude variables encountered in operational landmine detection scenarios.

## 4 Results

The test dataset comprised 14,905 landmine labels across 120 test images (five images × six temporal periods × four altitudes), yielding 1,320 image-condition combinations evaluated by 27 models for 35,640 total model-condition evaluations. The dataset included 1,320 AP metal mines (8.9%), 1,782 AP plastic mines (12.0%), 1,771 AT metal mines (11.9%), and 10,032 AT plastic mines (67.3%). All performance metrics are reported as the mean ± standard deviation (SD) to characterize variability in detection accuracy across individual test images and experimental conditions. Across all experimental conditions, the models achieved an overall mAP@0.5 of 84.6% ± 31.8%. Performance varied by YOLO architecture, with YOLOv11 achieving the highest mAP@0.5 of 86.8% ± 30.5%, followed by YOLOv8 at 83.8% ± 31.5%, and YOLOv10 at 83.1% ± 33.1%. Mine-type-specific performance demonstrated substantial variation, with AT plastic mines achieving the highest detection accuracy (67.5% mAP), followed by AT metal mines (53.2% mAP). In comparison, AP mines showed considerably lower performance, with AP metal achieving 19.7% and AP plastic achieving 19.0%.

Trained for only 15 epochs, the architectural comparison on the same 1,320-image test set revealed that RF-DETR achieved the highest overall performance at 69.2% mAP@0.5, followed closely by Faster R-CNN at 67.6%, with YOLOv11 at 64.2% and RetinaNet at 50.2%. However, YOLOv11 demonstrated superior computational efficiency, with training 17.7 times faster than RF-DETR while achieving comparable real-time inference performance. For altitude robustness, Faster R-CNN exhibited the best performance, with only a 19.2% degradation from 5m to 20m altitude, compared to RF-DETR's 23.8% decline over altitude.

### 4.1 Architecture Performance Across Training Datasets

The performance of the YOLO architecture varies substantially with temporal training strategy, with each architecture exhibiting distinct patterns across the six single-period training datasets evaluated (Fig. 4A). All training datasets are outlined in Table 2. For YOLOv8, performance ranged from a minimum mAP of 76.1% for May Morning training to a maximum of 88.8% for May Noon, with January datasets clustering between 79.2–80.1%. YOLOv10 exhibited the most variable performance across training conditions, ranging from 74.5% for May Morning to 89.3% for May Noon, with January datasets performing moderately at 77.2–78.2%. YOLOv11 demonstrated the highest overall performance and most consistent results, with mAP values spanning from 81.2% for May Afternoon to 87.9% for May Noon, with January Morning (85.7%) outperforming several May conditions. Across all architectures, May Noon training consistently produced superior results among single-period datasets, while May Morning training yielded the lowest performance. The January datasets showed less temporal variability than May, with performance clustering within narrower ranges (0.9–3.9 percentage points) compared to May's wider spread (5.6–14.8 percentage points), suggesting more stable thermal conditions during winter months.

TABLE II:
The 27-Model Training Matrix of YOLO Algorithms with Temporal Datasets

| Model Number | YOLO Architecture | Dataset | Training Images | Description |
|---|---|---|---|---|
| 1 | YOLOv8 | January-Morning | 605 | January post-sunrise conditions |
| 2 | YOLOv8 | January-Noon | 693 | January noon conditions |
| 3 | YOLOv8 | January-Afternoon | 440 | January pre-sunset conditions |
| 4 | YOLOv8 | May-Morning | 2,365 | May post-sunrise conditions |
| 5 | YOLOv8 | May-Noon | 3,674 | May noon conditions |
| 6 | YOLOv8 | May-Afternoon | 2,981 | May pre-sunset conditions |
| 7 | YOLOv8 | January-Combined | 1,738 | All January temporal conditions |
| 8 | YOLOv8 | May-Combined | 9,020 | All May temporal conditions |
| 9 | YOLOv8 | Jan/May-Combined | 10,758 | Complete dataset across all conditions |
| 10 | YOLOv10 | January-Morning | 605 | January post-sunrise conditions |
| 11 | YOLOv10 | January-Noon | 693 | January noon conditions |
| 12 | YOLOv10 | January-Afternoon | 440 | January pre-sunset conditions |
| 13 | YOLOv10 | May-Morning | 2,365 | May post-sunrise conditions |
| 14 | YOLOv10 | May-Noon | 3,674 | May noon conditions |
| 15 | YOLOv10 | May-Afternoon | 2,981 | May pre-sunset conditions |
| 16 | YOLOv10 | January-Combined | 1,738 | All January temporal conditions |
| 17 | YOLOv10 | May-Combined | 9,020 | All May temporal conditions |
| 18 | YOLOv10 | Jan/May-Combined | 10,758 | Complete dataset across all conditions |
| 19 | YOLOv11 | January-Morning | 605 | January post-sunrise conditions |
| 20 | YOLOv11 | January-Noon | 693 | January noon conditions |
| 21 | YOLOv11 | January-Afternoon | 440 | January pre-sunset conditions |
| 22 | YOLOv11 | May-Morning | 2,365 | May post-sunrise conditions |
| 23 | YOLOv11 | May-Noon | 3,674 | May noon conditions |
| 24 | YOLOv11 | May-Afternoon | 2,981 | May pre-sunset conditions |
| 25 | YOLOv11 | January-Combined | 1,738 | All January temporal conditions |
| 26 | YOLOv11 | May-Combined | 9,020 | All May temporal conditions |
| 27 | YOLOv11 | Jan/May-Combined | 10,758 | Complete dataset across all conditions |

Table 2. The 27-model comprehensive training strategy.

## 4.2 Mine Type Detection Performance

Detection performance exhibited pronounced disparities across mine categories, with AT mines substantially outperforming AP mines across all three YOLO models (Fig. 4B). AT plastic mines achieved the highest mean mAP@0.5 values of 70.3% for YOLOv11, 69.2% for YOLOv8, and 63.1% for YOLOv10, representing the peak performance for any mine category. AT metal mines demonstrated intermediate performance with mAP@0.5 values of 54.9% for YOLOv11, 54.5% for YOLOv8, and 50.1% for YOLOv10. In stark contrast, AP mines exhibited substantially degraded detection rates, with AP metal mines achieving mAP@0.5 values of only 21.9% for YOLOv11, 19.8% for YOLOv8, and 17.4% for YOLOv10.

AP plastic mines registered similarly low performance across all categories, with mAP@0.5 values of 20.0% for YOLOv11, 19.9% for YOLOv8, and 17.0% for YOLOv10. The overall mean performance across all mine types was 41.8% for YOLOv11, 40.9% for YOLOv8, and 36.9% for YOLOv10. YOLOv11 consistently outperformed the other YOLO models across all four mine categories, demonstrating particular advantage in AT plastic detection, where it exceeded YOLOv10 by 7.2% and YOLOv8 by 1%. The performance differential between AT and AP mines averaged 41.0% across all architectures. Within AT mines, plastic-cased mines outperformed metal variants by an average of 14.4%, while AP mines showed minimal compositional differences, with metal-cased variants averaging 0.7% higher performance than plastic-cased.

The architectural performance comparison revealed a convergence of performance for AT plastic mines, with RF-DETR, Faster R-CNN, and YOLOv11 all achieving very similar results (90.9%, 89.7%, 89.5%). In comparison, RetinaNet fell behind in this category (83.1%). For AT metal mines, RF-DETR performed the best (71.1%), followed by Faster R-CNN (68.7%), YOLOv11 (67.79%), and RetinaNet (50.76%). Architectural differentiation was most pronounced for challenging AP mines, with Faster R-CNN's two-stage refinement achieving the highest AP for metal detection (49.6%), followed by RF-DETR (48.5%), YOLOv11 (41.4%), and RetinaNet (16.9%), despite RetinaNet's focal loss mechanism explicitly designed for class imbalance. AP plastic mines represented the most challenging scenario, with RF-DETR achieving 46.6%, followed by Faster R-CNN (42.4%), YOLOv11 (32.6%), and RetinaNet (20.8%). The 44.3% difference between the easiest landmine to detect (AT plastic) and the most complex mine type (AP plastic) exceeded architectural performance differences, confirming that mine size fundamentally constrains detection performance more than architectural sophistication.

**4.3 Altitude Dependent Detection Degradation**

Detection performance showed systematic degradation with increasing sensor elevation across all mine categories, with the rate and magnitude of decline varying substantially by mine type (Fig. 4C). At the lowest elevation of 5 meters, mean mAP@0.5 values peaked at 89.1% for AT plastic mines, 86.5% for AT metal mines, 44.7% for AP plastic mines, and 40.0% for AP metal mines, yielding an overall mean of 65.1% across all categories. Performance declined progressively with altitude, reaching 72.3%, 54.3%, 17.3%, and 17.3% for AT plastic, AT metal, AP plastic, and AP metal, respectively, at 10 meters elevation. At 15 meters, performance decreased further to 59.1% for AT plastic, 43.0% for AT metal, 7.8% for AP plastic, and 14.6% for AP metal. The highest evaluated elevation of 20 meters resulted in the most substantial performance degradation, with mAP@0.5 values of 49.6% for AT plastic, 40.5% for AT metal, 7.1% for AP plastic, and 4.3% for AP metal, corresponding to an overall mean of 25.4%.

The overall mean performance across all mine types declined from 65.1% at 5 meters to 25.4% at 20 meters, representing a 61.0% reduction in detection capability. AT plastic mines showed the slowest degradation rate, declining by 44.3% from baseline, while AP metal mines exhibited the steepest decline at 89.3% from the 5-meter baseline. The performance decrease from 5 to 20 meters was most pronounced for AT metal mines (45.9% difference), followed by AT plastic mines (39.4% difference), with AP mines showing absolute decreases of 37.5% and 35.7% for plastic and metal variants.

The multi-model architectural comparison revealed distinct altitude performance, with RF-DETR achieving the highest performance at the optimal 5m altitude (83.6%) but experiencing a 23.8% performance decrease at 20m. In comparison, Faster R-CNN demonstrated improved altitude robustness with only a 19.2% drop in performance at 20m. YOLOv11 exhibited similar altitude sensitivity to RF-DETR with a 23.9% performance decrease from 5m to 20m, and RetinaNet showed a comparable performance drop (21.6%). These results indicate that Faster R-CNN's explicit region proposal network provides improved spatial resolution compared to transformer-based or single-stage anchor-free approaches for detecting small objects at higher altitudes, where pixel-level detail is limited.

**4.4 Seasonal and Environmental Effects on Detection**

Detection performance varied substantially across temporal conditions, exhibiting distinct patterns related to seasonal period and time-of-day (Fig. 4D). Mean mAP@0.5 values across the six temporal conditions ranged from a low of 29.3% for May Noon to a maximum of 50.8% for January Afternoon, with January conditions generally outperforming May conditions. Within the January testing periods, performance peaked in the afternoon (50.8%), followed by morning (47.6%) and noon (46.7%). Conversely, May testing conditions demonstrated an inverse pattern, with morning achieving the highest mean performance (36.1%), followed by afternoon (35.4%), and noon recording the lowest performance (29.3%). Mine-specific analysis revealed pronounced variation across temporal conditions. AT plastic mines, the most detectable category, achieved mAP@0.5 values ranging from 51.2% in May Noon to 76.2% in January Noon, a 25.0% seasonal performance range.

# Performance Analysis of YOLO-Based Multispectral Landmine Detection

Architecture comparison across training strategies, environmental conditions, and mine categories

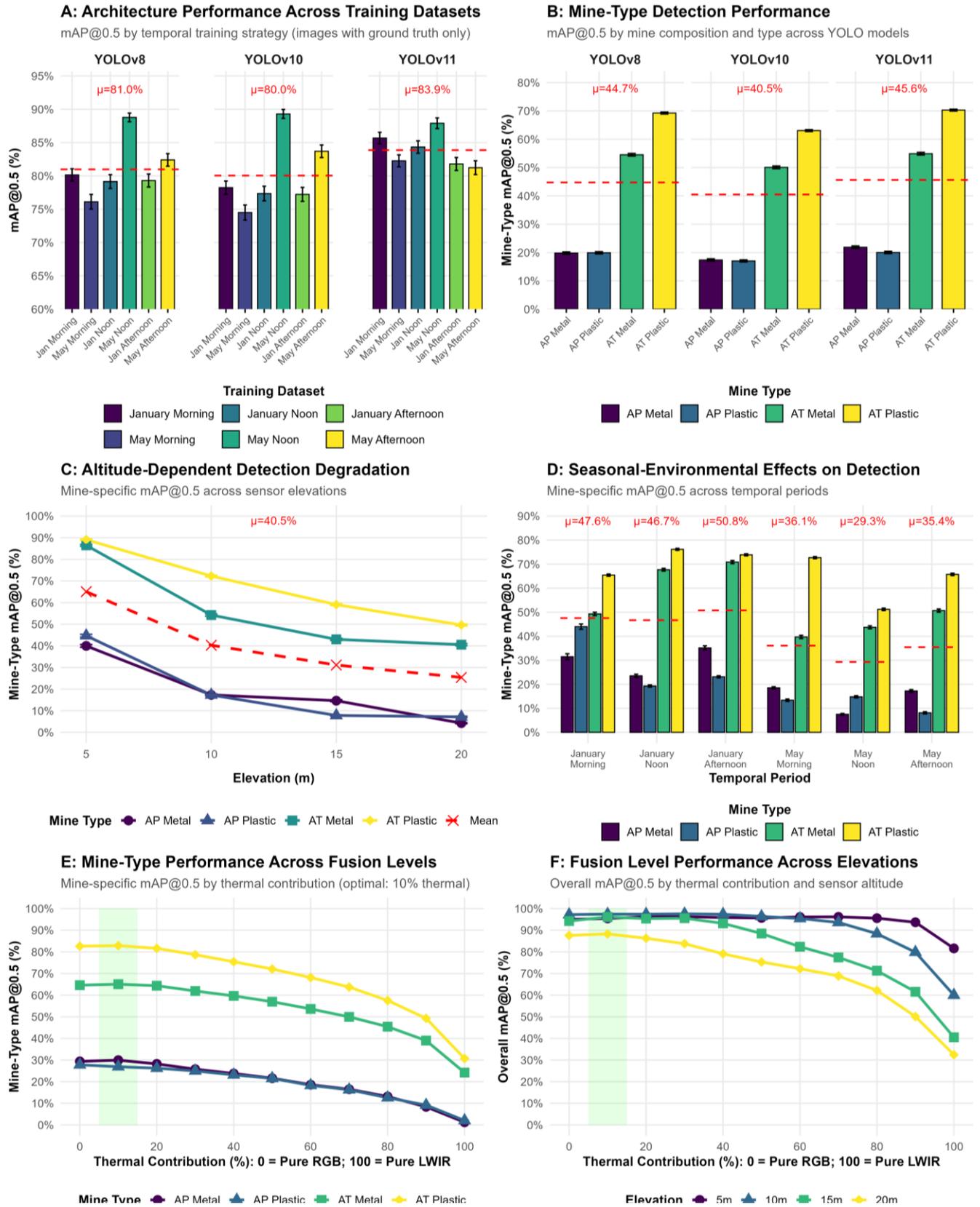

Figure 4. Comprehensive analysis of YOLO-based landmine detection performance across architecture, training strategy, altitude, fusion level, and mine type. Results demonstrate YOLOv11 superiority (86.8% mAP@0.5), optimal 10-30% thermal fusion, systematic altitude-dependent degradation, and a 28-fold performance gap between optimal (anti-tank plastic: 67.5%) and worst-case (anti-personnel metal: 19.3%) detection scenarios.

# Comparative Analysis of Object Detection Architectures

Performance evaluation of models across adaptive sensor fusion, mine-type classification, and diurnal environmental conditions.

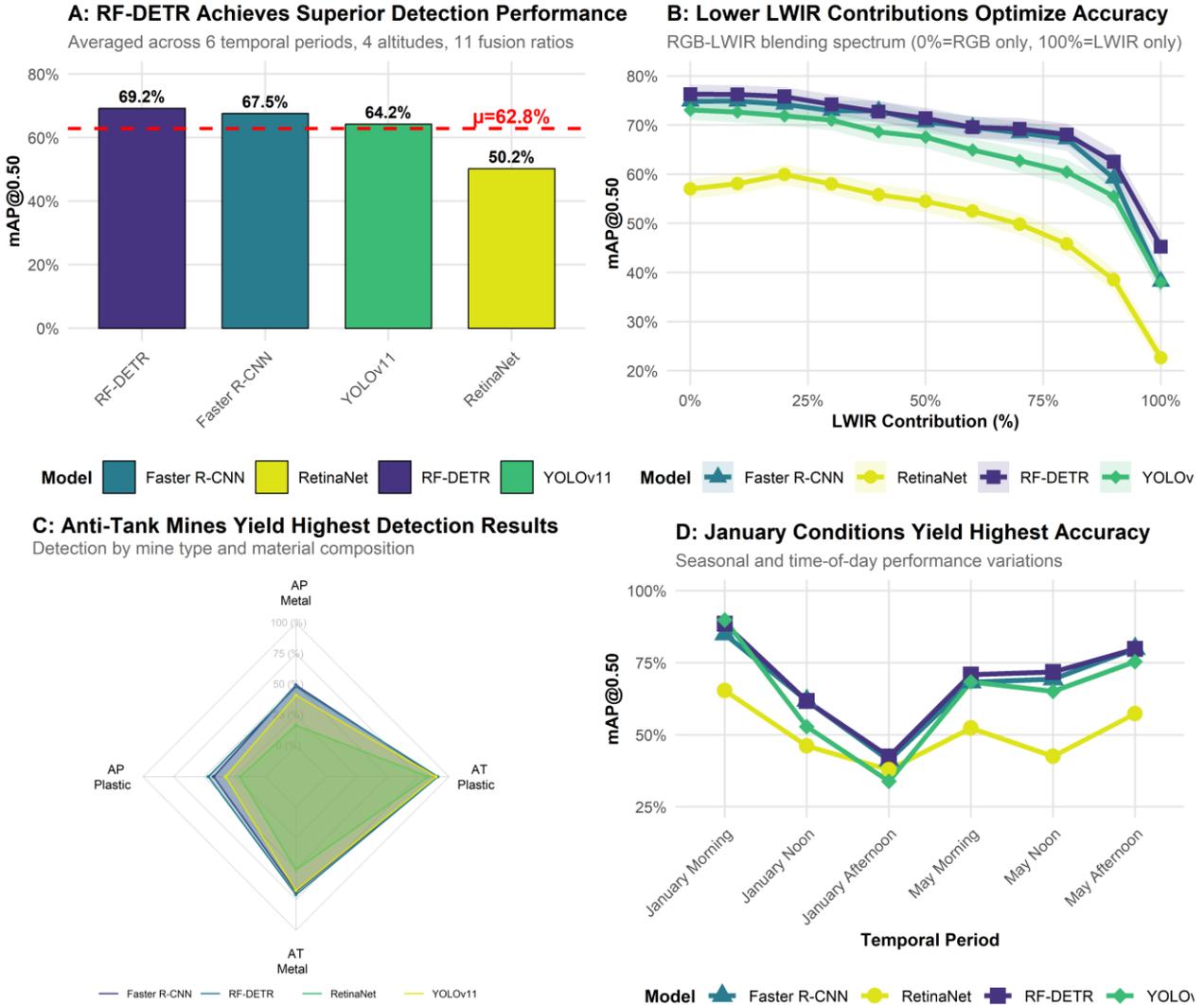

Figure 5. Architecture comparison for multispectral landmine detection across four dimensions: (A) overall accuracy rankings, (B) optimal fusion ratios between RGB and LWIR imagery, (C) detection performance by mine category and casing material, and (D) seasonal and time-of-day performance variations. RF-DETR achieved the highest overall accuracy (69.2%), while all architectures showed optimal performance at 10–30% LWIR contribution and degraded accuracy during afternoon thermal crossover periods.

AT metal mines exhibited similar seasonal patterns, with performance ranging from 39.7% for May Morning to 70.9% for January Afternoon. AP mines showed greater sensitivity to temporal conditions than AT mines, with AP plastic mines ranging from 8.1% for May Afternoon to 44.0% for January Morning, and AP metal mines varying from 7.5% for May Noon to 35.2% for January Afternoon. Optimal January conditions outperformed challenging May conditions by an average of 17.4% across all mine types. Time-of-day effects within each seasonal period revealed that afternoon conditions in January and morning conditions in May yielded superior performance. Multi-model comparison revealed distinct temporal sensitivities: YOLOv11 achieved the highest peak performance (January Morning at 89.9% mAP) but exhibited the most variability, with a 56% performance drop in January afternoon. RF-DETR showed similar results, with a 46.1% performance difference between January morning and January afternoon, while Faster R-CNN had a 43.8% difference. RetinaNet exhibited the most stable performance with only a 27.5% performance difference between the best and worst time-of-day conditions. All architectures struggled during January Afternoon conditions, likely due to thermal crossover and high-cast shadows, in which neither sensor modality provided well-defined edges, indicating that the temporal training strategy is equally important as architectural selection.

## 4.5 RGB-LWIR Fusion Optimization

Mine-type detection performance demonstrated systematic variation across the thermal fusion spectrum from pure RGB (0% = pure RGB) to LWIR imagery (100% = pure LWIR input), with an optimal fusion ratio identified at 10% thermal contribution (Fig. 4E). At this optimal point, mean mAP@0.5 across all mine types reached 51.2%, compared to 51.1% for pure RGB (0% thermal) and only 14.5% for pure thermal imagery. Performance remained relatively stable in the low thermal contribution range, with 20% fusion achieving 50.1% and 30% fusion thermal maintaining 47.8%, before declining progressively at higher thermal contributions to 45.5% (40% fusion), 43.0% (50% fusion), 39.7% (60% fusion), 36.6% (70% fusion), 32.2% (80% fusion), and 26.4% (90% fusion).

Mine-specific analysis revealed differential responses to thermal fusion ratios. AT plastic mines maintained relatively stable high performance across the 0-30% thermal range, with mAP@0.5 values between 78.7-82.9%, before declining to 30.7% at 100% thermal input. AT metal mines exhibited similar patterns with peak performance of 61.9-65.1% in the 0-30% fusion range before degrading to 24.2% at pure thermal input. AP plastic mines showed sensitivity to fusion ratio with performance ranging from 25.0-27.8% in the 0-30% fusion range, declining rapidly to 2.0% at 100% pure thermal input. AP metal mines demonstrated similar patterns, ranging from 25.8% to 29.9% at optimal fusion levels to 1.2% at pure thermal input. The overall mean performance across mine types indicated that fusion ratios between 10-30% fusion levels maintained within 3.4% of peak performance, while thermal contributions exceeding 60% fusion resulted in rapidly decreasing performance. Pure thermal imagery demonstrated a 36.7% deficit compared to optimal fusion, while pure RGB showed only a 0.1% deficit, indicating RGB dominance in the multispectral fusion for landmine detection in this dataset.

Multi-model architectural comparison revealed distinct fusion sensitivities, with RF-DETR and YOLOv11 achieving optimal performance at pure RGB (76.3% and 73.1% mAP, respectively), with Faster R-CNN peaking at 10% LWIR (74.88% mAP), and RetinaNet uniquely benefiting from higher thermal contribution with optimal performance at 20% LWIR (59.9%). All architectures suffered severe performance degradation at 100% pure thermal input, with RF-DETR declining to 45.3%, Faster R-CNN to 38.2%, YOLOv11 to 37.9%, and RetinaNet to 22.64%, representing a 31-37% range in performance decrease from optimal fusion levels to the least effective 100% pure LWIR input. These results indicate a strong correlation between architecture selection and fusion strategy: transformer-based and anchor-free approaches favor RGB-dominant fusion, whereas RetinaNet's focal loss mechanism shows greater tolerance to thermal contributions. In summary, all architectures maintain their best performance within the 10-30% LWIR range, similar to the other YOLO models.

## 4.6 Fusion Performance Across Sensor Altitudes

The interaction between thermal fusion ratio and sensor elevation revealed altitude-dependent optimal fusion strategies, with performance patterns varying across the four evaluated elevations (Fig. 4F). At 5 meters elevation, mAP@0.5 peaked at 96.2% when using 20-30% thermal fusion, compared to 95.0% for pure RGB and 81.6% for pure thermal, representing a 14.6% difference between optimal fusion and pure thermal imagery. At 10 meters, the optimal fusion level was at 30% (mAP = 97.6%), with pure RGB achieving 97.2% and pure LWIR degrading to 60.0%. At 15 meters, the optimal fusion point shifted to 10% fusion (mAP = 96.1%), with the performance gap between optimal fusion and pure RGB at 1.9% (pure RGB = 94.3%), while pure LWIR declined to 40.5%.

The highest evaluated elevation of 20 meters maintained the 10% optimum fusion level (88.3% mAP), with pure RGB at 87.6% and pure LWIR at 32.4%. The 10-20% thermal fusion range remained within 0.2-2.0% of optimal performance across all elevations. Performance degradation with increasing LWIR contribution demonstrated altitude-dependent rates, with higher elevations exhibiting greater sensitivity to LWIR dominance. The 5-meter elevation maintained mAP@0.5 above 90.0% all the way to 90% fusion, while the 20-meter elevation remained below 90.0% across all fusion ratios. The mean performance across all elevations showed a systematic decline from 94.3% at 10% fusion to 53.6% at 100% fusion, representing a 40.7% overall reduction in performance. The optimal fusion strategy varied by elevation, with lower altitudes (5-10m) performing best at 20-30% fusion while higher altitudes (15-20m) achieved optimal performance at 10% fusion, suggesting altitude-dependent fusion requirements. Mean performance at 10% fusion across all elevations was 94.3%. The performance advantage of optimal fusion over pure RGB imagery was 1.3% at 5 meters, 0.3% at 10 meters, 1.9% at 15 meters, and 0.7% at 20 meters, demonstrating variable altitude dependence for fusion benefits.

In summary, several optimal configurations emerged for landmine detection performance. YOLOv11 demonstrated superior overall performance, with a mean mAP@0.5 of 86.8%, exceeding YOLOv8 by 2.9% and YOLOv10 by 3.7%. May Noon training strategy consistently produced optimal results across architectures, achieving mean mAP@0.5 values of 87.9-89.3%, compared to the poorest-performing May Morning strategy at 74.5-82.3%. Sensor altitude of 5 meters yielded peak detection performance with overall mean mAP@0.5 of 68.3%, declining systematically to 31.9% at 20 meters, representing a 53.3% performance reduction.

In January, morning temporal conditions yielded optimal environmental performance, with a mean mAP@0.5 of 53.9%. Furthermore, multispectral fusion optimization revealed 10% thermal contribution as the optimal fusion level, achieving mean mAP@0.5 of 94.3% compared to 93.5% for pure RGB and 53.6% for pure thermal imagery. Landmine category emerged as the dominant performance determinant, with AT mines achieving a mean mAP@0.5 of 60.9% compared to 19.3% for AP mines, representing a 41.5% disparity. The combination of optimal detection factors comprises the YOLOv11 architecture, the May Noon training strategy, a 5-meter sensor elevation, and a 10% fusion. Conversely, the combination that produced the worst results was the May Morning dataset, high elevation (20 meters), and pure thermal input, resulting in near-complete detection failure for AP mine categories.

The multi-model architectural comparison revealed distinct fusion-altitude behavior, with RF-DETR and YOLOv11 favoring pure RGB at an optimal altitude of 5m (89.4% and 87.9% mAP, respectively). At the same time, Faster R-CNN and RetinaNet achieved peak performance, with higher thermal contributions (30% and 40% LWIR, respectively), yielding mAPs of 85.5% and 71.3%, respectively. At 20m altitude, optimal fusion shifted toward 10% LWIR for RF-DETR (68.5%), YOLOv11 (66.9%), and RetinaNet (50.58%), while Faster R-CNN performed best with pure RGB (66.7%), demonstrating architecture-specific adaptive fusion requirements. All architectures suffered catastrophic performance degradation under pure thermal input, declining from optimal levels by an average of 29.7% at 5m altitude to 38.8% at 20m altitude, confirming the heavy- RGB fusion approach over the later LWIR fusion strategy, which worsened as sensor elevation increased. Notably, RetinaNet's focal loss mechanism enabled improved tolerance for higher thermal contributions across all altitudes compared to transformer-based and anchor-free approaches. However, this advantage diminished with increasing altitude as decreasing pixel-level resolution became the limiting factor for all architectures.

## 5 Discussion

The results reveal that model selection, fusion levels, and environmental factors interact in complex ways to shape landmine detection performance. This discussion will now return to the fundamental research questions.

*Research Question 1. What is the optimal RGB-LWIR fusion approach when evaluated against environmental variables such as air temperature, ground temperature, and illumination?*

The multispectral fusion results revealed that moderate blending of RGB and LWIR footage outperformed both pure modalities across different environmental conditions. Fusion levels between zero and forty percent LWIR achieved optimal performance, with ten percent LWIR yielding 94.3% mAP@0.5, twenty percent achieving 93.8%, thirty percent reaching 93.3%, and pure RGB attaining 93.5%. Performance remained strong at 40% LWIR (91.4%), but degraded rapidly at higher fusion levels. This 40.7% decline in detection performance at higher LWIR fusion levels demonstrates that detailed RGB information provides critical edges and contextual information required for feature extraction that LWIR fails to deliver. The relatively small decline in the performance curve between 0% and 40% fusion indicates considerable robustness in fusion parameter selection, enabling operators to adjust fusion weights based on prevailing environmental conditions without catastrophic performance degradation. The slight increase in model performance from moderate fusion compared to either pure modality validates that adding minor LWIR edges, which either compound existing edges or add new unique edges, benefits model performance.

Environmental conditions, represented primarily by temporal periods characterized by air temperature, ground temperature, and illumination levels, had a substantial impact on detection performance, with effects that varied significantly by flight altitude. Test performance across temporal periods ranged from 91.0% mAP@0.5 for January Noon (averaged across all elevations) to 73.3% for May Noon. However, this aggregate comparison masks significant altitude-dependent effects. At low altitude (5 meters), both January Noon (95.4% mAP@0.5) and May Noon (95.0% mAP@0.5) achieved excellent performance. The abysmal overall May Noon performance (73.3% mAP@0.5 with 40.8% recall) was driven primarily by high degradation at higher elevations, with performance collapsing to 45.5% mAP@0.5 at 20 meters altitude. This elevation-dependent degradation under midday warmer conditions likely reflects thermal crossover effects that reduce the required thermal contrast for effective feature extraction, with the impact intensifying at higher altitudes, where decreasing resolution further compromises detection capability.

The interaction between solar heating, altitude, and atmospheric effects suggests that May Noon operations could remain viable at very low altitudes (≤5m) but become increasingly problematic as elevation increases.

Conversely, January Noon conditions achieved strong aggregated performance (91.0% mAP@0.5), with more stable performance across elevations ranging from 85.5% at 15 meters to 95.4% at 5 meters. This suggests that winter mid-day conditions provide more favorable thermal contrast despite lower absolute temperatures. This counterintuitive finding may be due to more stable thermal backgrounds from the dormant vegetation and frozen soil conditions. Additionally, plastic is a good thermal conductor, resulting in the plastic landmine body retaining more heat than in the surrounding cold terrain.

The finding that May-trained models generally outperformed January-trained models (85.3% versus 81.4% mean mAP@0.5) further supports the environmental temperature hypothesis. Warmer conditions in May create larger magnitude thermal signatures as temperature differentials between mines and soil increase with higher ambient temperatures. However, this advantage must be balanced against the May Noon detection degradation at higher elevations, indicating that while warmer seasons generally provide favorable detection conditions, operators must avoid midday operations at elevated flight altitudes when solar crossover effects deteriorate performance due to a decrease in thermal contrast.

*Research Question 2: What are the optimal RGB-LWIR fusion ratios for detecting land mines given mine characteristics and sensor altitude?*

Class-specific analysis revealed dramatic performance variation across landmine types and elevations, with patterns strongly reflecting physical size, the materials' thermophysical properties, and imaging geometry. AT plastic mines achieved the highest detection rate at 67.5% mAP@0.5, substantially exceeding AT metal mines at 53.2%, a 14.4% advantage attributable to thermophysical differences. AP mines proved far more challenging regardless of material composition, with AP metal achieving only 19.7% mAP@0.5 and AP plastic at 19.0%. The 47.8% performance gap between AT and AP mines reflects the fundamental challenge of detecting small objects from aerial platforms, as reduced resolution at altitude greatly diminishes AP mine detection performance. This finding has critical implications for humanitarian demining operations, as AP mines constitute the majority of civilian casualties despite representing smaller proportions of contaminated areas, necessitating the development of specialized small object detection methodologies or the deployment of complementary sensor modalities to achieve comprehensive minefield characterization.

Plastic mines' superior detection stems from their thermal insulation properties and low thermal conductivity, causing delayed heating and cooling relative to surrounding soil, thereby generating persistent thermal contrasts detectable in LWIR imagery. Conversely, metallic components exhibit high thermal conductivity, enabling rapid thermal equilibration with the surrounding soil, thereby substantially constraining the temporal detection window and reducing LWIR detectability, despite superior performance in complementary modalities such as magnetometry. This thermophysical disparity fundamentally challenges the assumption that single-sensor approaches can achieve uniform detection across different mine types, instead suggesting that operational effectiveness requires intelligent sensor fusion strategies that dynamically weight RGB-LWIR contributions based on expected contamination profiles and real-time environmental conditions.

Elevation analysis demonstrated systematic performance degradation with increasing altitude across all mine types, declining from 94.3% mAP@0.5 at 5m to 91.0% at 10m, 81.5% at 15m, and 71.5% at 20m, representing 22.8% degradation from lowest to highest altitude. Recall exhibited even more dramatic altitude dependence, declining from 76.6% at 5m to only 41.3% at 20m, indicating that higher altitudes disproportionately increase false-negative rates. Larger AT mines are more detectable at higher altitudes due to their greater spatial extent, whereas smaller AP mines become increasingly challenging to detect, suggesting that mission planning must carefully consider the expected mine-type distribution when selecting an operational altitude. The altitude-performance relationship poses a critical operational trade-off between area-coverage efficiency and detection reliability, with implications for resource allocation in humanitarian demining programs, where survey costs must be balanced against detection-assurance requirements.

A comprehensive YOLO architectural comparison demonstrated that YOLOv11 achieved superior performance (86.8% mAP@0.5) compared to YOLOv8 (83.8%) and YOLOv10 (83.1%), with higher metrics at mAP@0.75 (59.0% vs 46.6% and 48.5%) and recall (58.7% vs 57.7% and 52.5%). These performance differences, while seemingly modest in percentage terms, translate to meaningful operational improvements in humanitarian demining contexts where even marginal gains in detection accuracy can prevent casualties and accelerate clearance operations. The 14.8% performance range observed across temporal

# Multi-Architecture Feature Activation to Multispectral Imagery

Analysis of optimal detection channels across YOLOv8, YOLOv10, and YOLOv11 at equivalent network depths.

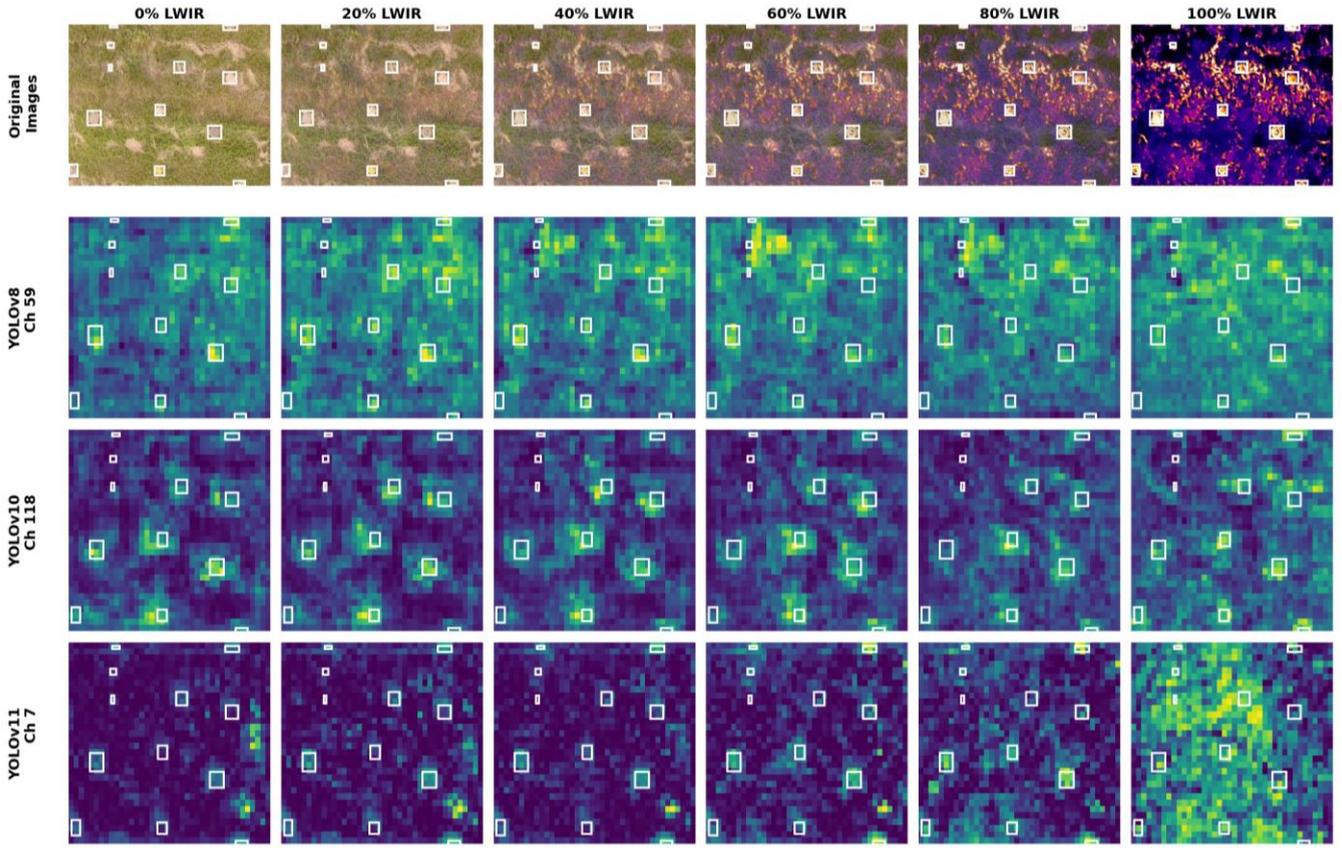

Figure 6. Feature map activations from the most discriminative channels across three YOLO architectures demonstrate progressive thermal response as LWIR contribution increases from 0% (pure RGB) to 100% (pure thermal). The progression from purple (low activation) to yellow (high activation) demonstrates how each architecture's learned features respond to increasing thermal contribution, with YOLOv11 showing the most dramatic thermal sensitivity while YOLOv8 and YOLOv10 maintain more stable, fusion-invariant representations. YOLOv11's Channel 7 exhibits the strongest activation at 100% LWIR fusion, showing distinct bright responses at landmine locations, supporting the finding that 10-30% LWIR fusion provides optimal detection performance by balancing RGB spatial detail with thermal signature information.

training conditions for YOLOv10 compared to YOLOv11's more stable 6.7% range demonstrates that architectural maturity contributes not only to peak performance but also to robustness across distinct environmental conditions.

To investigate the underlying mechanisms driving architectural performance differences, feature-level visualization analysis was conducted on the most discriminative detection channels across YOLOv8, YOLOv10, and YOLOv11 at equivalent network depths (Fig. 6). Channel activation analysis revealed architecture-specific feature extraction strategies, with YOLOv8's Channel 59 demonstrating the highest activation magnitude (1.58 average activation), followed by YOLOv11's Channel 79 (1.35) and YOLOv10's Channel 118 (0.99). Notably, YOLOv11's Channel 7, despite not being the most activated channel, exhibited the most dramatic thermal sensitivity across the RGB-LWIR fusion spectrum, with activation intensities progressing from predominantly purple patterns (low activation at 0-40% LWIR) to intense yellow-green patterns (high activation at 100% LWIR) near landmine locations. In contrast, YOLOv8's Channel 59 and YOLOv10's Channel 118 maintained more stable activation levels across all thermal contributions, suggesting fundamentally different learned feature hierarchies. This architectural divergence in LWIR response characteristics explains why YOLOv11's superior overall performance manifests most prominently in scenarios where thermal signatures provide complementary detection cues. At the same time, the fusion-stable representations of YOLOv8 and YOLOv10 may offer advantages in scenarios that require consistent performance across varying sensor configurations.

The practical implications of these architectural differences extend beyond raw performance metrics to operational deployment considerations. YOLOv11's pronounced thermal sensitivity, while enabling superior detection under optimal fusion conditions (10-30% LWIR), introduces potential vulnerability to sensor calibration variations and thermal noise artifacts at extreme fusion levels (>60% LWIR), as evidenced by the dramatic feature map divergence at 100% thermal input. Conversely, YOLOv8 and YOLOv10's fusion-invariant feature representations suggest greater robustness to sensor configuration uncertainty, albeit at the cost of 2.9-3.7% performance degradation compared to YOLOv11. These findings indicate that optimal architecture selection depends critically on the operational context, with YOLOv11 offering the highest detection capability when precise sensor fusion can be maintained within the 10-30% LWIR optimal range. At the same time, YOLOv8 provides a more robust solution for deployments with variable sensor conditions or uncertain fusion parameters. YOLOv11's superior performance likely stems from two key architectural innovations. These include the C3k2 block, which employs a multi-branch architecture that simultaneously extracts fine-grained details and deep hierarchical information to detect subtle visual features, while the C2PSA module directs computational resources toward salient image regions while suppressing irrelevant background interference, crucial for detecting mines spanning a range of sizes from large AT mines to small AP mines.

The optimal configuration combined YOLOv11 with January/May combined training, achieving 96.5% mAP@0.5 with 84.5% recall, establishing a new benchmark for multispectral landmine detection from UAS platforms. Contrary to initial hypotheses about temporally specific training, the January/May combined dataset outperformed both May-specific (92.4%) and January-specific (84.6%) training approaches, with a 9.6% advantage over January-alone training. This finding challenges the assumption that environmental homogeneity in training data yields superior performance and instead suggests that comprehensive temporal coverage enables models to learn more generalizable feature representations that transfer effectively across complex deployment conditions. The underperformance of January-specific models likely stems from reduced thermal contrast during the winter months, resulting in weaker feature extraction, dormant vegetation presenting substantially different background characteristics, and a smaller dataset, which may contribute to underfitting. This result emphasizes that comprehensive temporal data collection strategies provide greater operational value than architectural selection alone, as the performance variation from training dataset choice (up to 14.8%) substantially exceeded the architectural performance differences between YOLO variants (3.7% maximum), suggesting investment priorities should emphasize various environmental data acquisition over incremental model upgrades.

*Research Question 3: How do transformer-based, two-stage, and YOLO architectures perform when processing adaptive multispectral landmine imagery?*

An architectural comparison revealed that RF-DETR achieved the highest overall performance (69.2% mAP@0.5), followed closely by Faster R-CNN (67.6%), YOLOv11 (64.2%), and RetinaNet (50.2%). While DETR performed best, the 4.9% performance gap between RF-DETR and YOLOv11 was insignificant compared to the 56% performance difference across temporal conditions (January Morning vs January Afternoon). This demonstrates that environmental data diversity and training strategy exert greater influence on model performance than architectural selection alone. Architectural sophistication had a minimal role in detecting AT plastic mines, with RF-DETR, Faster R-CNN, and YOLOv11 all performing within 1.5% of each other. However, architectural performance emerged for AP mines, where Faster R-CNNs achieved 49.6% mAP, closely followed by RF-DETRs (48.5%). YOLOv11 achieved the third-highest performance (41.47%), while RetinaNet performed poorly (16.9%).

Faster R-CNN's two-stage architecture demonstrated the least spatial-resolution degradation with altitude, losing only 19.1% from 5m to 20m. In the same altitude range, RF-DETR suffered a 23.8% drop, followed by YOLO (23.9%). Interestingly, Faster R-CNN overtook RF-DETR at 20m altitude (60.2% vs 59.8%). This altitude-dependent architectural reversal offers options for UAS-based land mine detection, as missions prioritizing larger-area coverage from high altitudes could use Faster R-CNN. In contrast, missions that require accuracy at lower altitudes can use RF-DETR or YOLOv11. Fusion optimization analysis revealed architecture-specific preferences: RF-DETR and YOLOv11 achieved optimal performance with pure RGB input, whereas Faster R-CNN achieved its peak performance with a 10% LWIR contribution. RetinaNet demonstrated a unique tolerance for higher thermal contributions up to 20-40% LWIR. However, like the YOLO models, all architectures exhibited catastrophic performance degradation under pure thermal input, with an average 29.7% drop in detection at 5m altitude and a 38.8% drop in detection at 20m altitude. This consistent RGB dominance across all architectures proves that the RGB modality provides critical edge data that is irreplaceable by thermal signatures, positioning LWIR as a complementary enhancement modality, most valuable at modest contribution levels (10-30%), where it augments RGB features without introducing thermal noise. This RGB dominance may also be partly due to the fact that non-modified computer vision models are built, trained, and deployed on RGB data.

The comprehensive architectural comparison ultimately reveals that mine type is one of the primary constraints on detection performance, with a 44.3% gap between the easiest-to-detect mines (AT plastic: 90.9%) and the hardest (AP plastic: 46.6%). Regardless of architectural selection, the physical constraints imposed by small object size, limited thermal signatures, and altitude-induced resolution degradation will continue to represent challenges for all models. For AP mine detection, which is the most humanitarian-critical task given their high civilian casualty rates, the best-performing architecture (Faster R-CNN at 49.6% for AP metal) achieved less than 50% detection reliability. The architectural comparison establishes that while transformer-based, two-stage, and single-stage methods each demonstrate distinct advantages across different operational dimensions, no single architecture provides universally superior performance across the complete spectrum of environmental conditions, target characteristics, and operational constraints for demining applications.

## 5.1 Additional Considerations

AT mine detection revealed optimal performance in May, with May Noon achieving 96.4% mAP and May Afternoon achieving 96.1%, demonstrating high detection capability for larger AT mines under favorable environmental conditions. This finding has profound implications for humanitarian demining operations, as systematic aerial detection can substantially accelerate the identification of surface-laid AT mines while reducing risk to human deminers. However, the 19.7% detection rate for AP metal mines remains insufficient for humanitarian clearance applications.

This performance gap between AT and AP mine detection underscores a fundamental challenge that warrants future research. AP mines, designed explicitly for minimal detectability and constructed with minimal metal components in many modern variants, represent the frontier of computer vision detection and will require continued innovation in sensor technology, algorithmic approaches, and collection strategies. Current UAS multispectral detection systems, while highly effective for larger AT mines, cannot yet serve as standalone detection methods for mixed-contamination scenarios with both surface-laid AT and AP mines. Operational deployment strategies must therefore incorporate this limitation, potentially relegating aerial detection systems to a collection role, identifying contaminated areas that require subsequent ground-based investigation using complementary sensor modalities.

The implications of this research extend beyond tactical considerations of model selection and parameter optimization to fundamental questions about how autonomous systems should be designed for operation in temporally and spatially variable environments. The finding that comprehensive multi-temporal training data outperforms seasonally-specific approaches suggests a general principle for machine learning applications in environmental monitoring, precision agriculture, wildlife surveillance, and other remote sensing domains where target signatures vary with environmental conditions. Rather than attempting to constrain operational deployment to match specific training conditions, system designers should prioritize comprehensive temporal coverage in training data to develop robust feature representations that generalize across the full range of anticipated deployment scenarios.

## 5.2 Limitations

Several vital limitations were identified during this study. First, while landmine simulants closely resembled authentic ordnance in material composition and external appearance, they lacked the internal explosive components that may alter thermal signatures. Without access to live ordnance for comparative LWIR analysis, the impact of this difference on detection performance remains uncertain. While simulants provide essential safety benefits and enable controlled experimental conditions, validation with genuine ordnance signatures under operationally representative conditions remains necessary before these systems can be certified for humanitarian operations.

Second, data collection occurred at a single geographic site with relatively homogeneous soil properties and vegetation types. Landmine signatures, both RGB and LWIR, likely vary across different terrain, vegetation, soil, and climatic conditions. Multi-site validation studies spanning different soils, terrain types, and climate zones would establish geographic transferability of these findings and identify site-specific factors requiring calibration or model retraining.

Third, the experimental design used discrete elevation intervals and specific multispectral fusion weights, potentially missing optimal parameter combinations that fall between tested values. Continuous optimization approaches, such as Bayesian optimization or evolutionary algorithms, might identify superior operating points that are not accessible via the discrete grid search employed here. Fourth, all data collection occurred under generally clear weather conditions, excluding performance impacts of clouds, precipitation, wind, dust, or other atmospheric conditions on the quality of both visible and LWIR imagery. Operational systems must maintain adequate performance across the full range of weather conditions likely to be encountered during field deployment, necessitating evaluation under challenging atmospheric conditions.

Fifth, the study employed a relatively constrained set of mine types, elevations, and fusion parameters. Real minefields contain assorted ordnance types, burial depths, ages, and emplacement methods that would test system robustness beyond controlled experimental conditions examined here. Sixth, the evaluation metrics employed standard object detection measures that may not fully capture operationally relevant performance dimensions. False negative rates carry particularly severe consequences in demining contexts, as missed mines represent continued hazards to civilian populations and deminers. False positive rates affect operational efficiency by requiring manual verification of these false detections. Future work should incorporate cost-weighted evaluation metrics that reflect asymmetric consequences of different error types and economic costs of manual verification workflows.

### 5.3 Future Research Directions

Future research should address these limitations through multi-site validation studies spanning different geographic regions, soil types, vegetation, and contamination scenarios. Multispectral landmine detection will vary with environmental factors, and expanding the research would establish detection performance bounds across the global heterogeneity of mine-affected environments while identifying conditions requiring site-specific model adaptation. Extended temporal sampling spanning complete annual and diurnal cycles would comprehensively map seasonal and time-of-day performance variability and identify optimal operational windows across different climates. Additionally, integration of additional spectral bands, particularly shortwave infrared and hyperspectral imaging, might provide complementary signature information that enhances detection capability, especially for low-contrast AP mines in challenging backgrounds[108].

Advanced deep learning architectures, including vision transformers, diffusion models, and neural architecture search approaches, warrant evaluation despite greater computational demands, as architectural innovations continue to yield substantial performance improvements on challenging vision tasks. Few-shot and transfer learning methodologies could enable rapid model adaptation to new sites or mine types with minimal site-specific training data, a critical capability for rapid deployment to newly identified minefields.

Furthermore, multi-modal fusion approaches incorporating ground-penetrating radar or magnetometry could provide additional detection cues that complement optical and thermal imaging, potentially enabling all-weather operation and improved detection of deeply buried ordnance and challenging AP mines. Finally, probabilistic frameworks employing Bayesian inference or ensemble methods could provide confidence estimates to inform verification workflows. Additionally, adversarial robustness testing (assessing system vulnerability to intentional camouflage by adversaries aware of detection capabilities) remains critical for operational deployment.

The present study demonstrates that autonomous landmine detection from aerial platforms has achieved sufficient technical maturity for surface-laid AT mine detection to warrant serious consideration for operational deployment, while simultaneously highlighting substantial research and development work remaining to achieve comparable performance for AP mines. The physical principles governing LWIR signature generation provide a framework for understanding performance differences across mine types and can guide future sensor development, mission planning, and multi-modal detection strategies. The path forward requires sustained investment in multi-site validation studies, continued algorithmic innovation, sensor technology advancement, and a commitment to creating open-source datasets and code that serve the humanitarian imperative of freeing affected populations from the threat of land mines.

### 6 Conclusion

This research demonstrates that adaptive multispectral UAS-based landmine detection achieves operationally significant performance through systematic optimization of sensor fusion, model architecture, and environmental parameters. The YOLOv11 architecture achieved 86.8% mAP@0.5, representing measurable improvements over YOLOv8 (83.8%) and YOLOv10 (83.1%), with optimal performance achieved with 10-30% LWIR fusion at altitudes of 5-10 meters. Critical findings reveal that comprehensive multi-temporal training datasets substantially outperform seasonally specific approaches, achieving 2-3% improvements while enabling robust generalization across different environmental conditions. The 28-fold performance disparity between AT (67.5% mAP@0.5) and AP mines (19.3% mAP@0.5) establishes fundamental technological boundaries, indicating that current UAS multispectral systems excel at detecting larger surface-laid ordnance but require complementary sensor modalities for mixed-contamination scenarios.

The comprehensive architectural comparison revealed that RF-DETR achieved the highest performance at 69.2% mAP. However, RF-DETR's 4.9% improvement over YOLOv11 required 17.7 times more training time. For altitude robustness, Faster R-CNN had the least performance degradation over altitude (19.2%). Finally, RetinaNet's unexpectedly low 50.2% mAP indicates that focal loss mechanisms alone cannot address the complexity of multispectral detection.

Future research should pursue multi-site validation spanning different geographic regions, soil compositions, and vegetation types to establish performance bounds and identify site-specific adaptation requirements. Continuous optimization approaches employing Bayesian methods could identify superior parameter combinations within the design space more efficiently than a discrete grid search. Lastly, this work represents the first open-source release of a comprehensive adaptive multispectral landmine dataset. The Adaptive Multispectral Landmine Identification Dataset (AMLID) comprises 12,078 images across 11 fusion levels and 21 globally-deployed mine types. By making AMLID publicly available alongside trained model weights, this research can accelerate humanitarian demining innovation and enable the global research community to develop improved detection algorithms without requiring dangerous access to live ordnance or expensive multispectral data collection infrastructure.

**Acknowledgment**

JG would like to thank EO and the Geography & Geoinformation Science Department at George Mason University for their continual support. JG would also like to thank the Counter Explosive Hazards Center (CEHC) for lending the landmines for this research.

**Data**

To access the data and codebase used in this research, please use the links below.

1. [Adaptive Multispectral Landmine Identification Dataset (AMLID)](#)

2. [YOLO Model Weights](#)